\documentclass[11pt]{article}

\usepackage[preprint]{acl}

\usepackage{times}
\usepackage{latexsym}
\usepackage[T1]{fontenc}
\usepackage[utf8]{inputenc}
\usepackage{microtype}
\usepackage{inconsolata}
\usepackage{graphicx}
\usepackage{booktabs}
\usepackage{amsmath}
\usepackage{tikz}
\usetikzlibrary{arrows.meta, positioning, fit, backgrounds}

\title{Dissecting Agentic RAG: A Component Ablation\\
for Multi-Hop QA with a Local 7B Model}

\author{Sheroz Shaikh \\
  Independent Researcher \\
  \texttt{shaikh.sheroz07@gmail.com}}

\begin{document}
\maketitle

\begin{abstract}
Agentic retrieval-augmented generation (RAG) systems combine iterative reasoning loops, query decomposition, and adaptive retrieval to tackle multi-hop question answering. However, the contribution of each component remains poorly understood, particularly under resource-constrained settings using only local language models. Many agentic designs add adaptive retrieval routing and deeper retrieval loops on the assumption that the added complexity helps. To test whether it does, we run a controlled ablation study of a full agentic RAG pipeline evaluated on 5,000 questions from the HotpotQA distractor development set \citep{yang2018hotpotqa} using a local 7B parameter model (Qwen2.5-7B-Instruct; \citealt{qwen2025}). Our full pipeline achieves EM=53.2\% and F1=61.6\%, compared to a single-pass dense-retrieval baseline of EM=43.1\% and F1=54.0\%. Across eight ablation conditions, we find that: (1) fixed hybrid retrieval via reciprocal rank fusion consistently outperforms rule-based adaptive routing (+1.8 EM, +1.9 F1), as the routing heuristic over-routes to BM25 by firing on named entities present in nearly all multi-hop sub-questions; (2) two retrieval iterations over the decomposed sub-questions capture 95\% of the gains of five, with no meaningful benefit from deeper loops; and (3) query decomposition and cross-encoder reranking each contribute statistically significant but smaller gains ($p<0.01$ and $p<0.001$ respectively). Taken together, on a fixed local-model budget the simpler and fixed choices turn out to be competitive with or better than their adaptive versions: most of the gain comes from running a short retrieval loop, not from adaptive routing or from many iterations. We use no proprietary APIs or large-scale compute.
\end{abstract}

\section{Introduction}

Multi-hop question answering (QA) requires reasoning across multiple documents, chaining intermediate facts before reaching a final answer \citep{yang2018hotpotqa}. Standard retrieval-augmented generation (RAG) \citep{lewis2020rag}, which performs a single retrieval pass before synthesis, struggles in this setting because one retrieval step rarely surfaces the full evidence chain needed for complex questions.

Agentic RAG addresses this limitation by embedding retrieval inside an iterative reasoning loop. Drawing on the ReAct framework \citep{yao2023react}, these pipelines allow a language model to alternate between reasoning and retrieval, refining its evidence gathering based on what it has already found. Query decomposition strategies \citep{press2022selfask} and adaptive retrieval routing \citep{jiang2023flare} further augment these pipelines, in principle tailoring retrieval behavior to each query's characteristics.

Although agentic RAG is now widely adopted, these systems are complex, and the relative contribution of each component is not well understood. Existing work largely demonstrates end-to-end gains on multi-hop benchmarks \citep{trivedi2022interleaved} but rarely isolates individual components in a controlled manner. Ablation studies that do exist typically use large proprietary models, making it unclear whether the findings transfer to resource-constrained deployments.

We address this gap with a systematic component-level ablation of an agentic RAG pipeline, run entirely on a local Qwen2.5-7B-Instruct model with no proprietary API calls. We run eight controlled ablation conditions on 5,000 HotpotQA distractor questions, varying the retrieval step depth, the presence of query decomposition, the reranking stage, and the retrieval strategy. Our aim is empirical rather than a new model. Using these ablations and paired significance tests, we ask whether each added component actually helps on a local 7B model. For two of them the answer is no: deeper retrieval loops stop helping after about two steps, and the rule-based router does worse than simply using fixed hybrid retrieval.

Our main contributions are:
\begin{itemize}
    \item A controlled component ablation of a full agentic RAG pipeline on a single local 7B model (Qwen2.5-7B-Instruct), with paired significance tests and bootstrap confidence intervals over 5{,}000 HotpotQA questions.
    \item Fixed hybrid retrieval (reciprocal rank fusion over dense and sparse signals) does better than the rule-based adaptive router meant to replace it, by +1.8 EM and +1.9 F1 ($p<0.001$). The router fires on named entities, which appear in almost every multi-hop sub-question, so it sends most queries to BM25 and loses the complementary dense signal. The per-strategy scores that seem to favor routing appear to be a selection-bias effect (Section~\ref{sec:analysis}).
    \item Two retrieval iterations capture 95\% of the gain from five ($p<0.001$ vs.\ one step), and deeper loops add nothing measurable. The large improvement over the baseline comes from running the loop at all, not from running it many times.
    \item Query decomposition ($p=0.004$) and cross-encoder reranking ($p<0.001$) each contribute statistically significant but smaller improvements. Reranking is retained unconditionally; decomposition improves accuracy but, unlike reranking, carries a non-trivial latency cost.
\end{itemize}

\section{Related Work}

\paragraph{Retrieval-Augmented Generation.}
\citet{lewis2020rag} introduced RAG, pairing a dense retriever with a generative model for knowledge-intensive NLP tasks. Dense passage retrieval \citep{karpukhin2020dpr} and late-interaction models \citep{khattab2020colbert} advanced the retrieval component. Hybrid retrieval via reciprocal rank fusion \citep{cormack2009rrf}, combining dense and sparse \citep{robertson2009bm25} rankings, has shown consistent gains across tasks and serves as a key baseline in our study.

\paragraph{Iterative and Agentic Retrieval.}
IRCoT \citep{trivedi2022interleaved} interleaves chain-of-thought reasoning steps with retrieval, showing that iterative evidence accumulation substantially improves multi-hop QA. Self-Ask \citep{press2022selfask} decomposes complex questions into answerable sub-questions before retrieval. ReAct \citep{yao2023react} provides a general framework for language model agents that alternate between reasoning and action. We adopt a plan-and-execute variant of this idea, in which sub-questions are generated up front by the decomposer and then retrieved iteratively, rather than interleaving a fresh reasoning step before every retrieval. FLARE \citep{jiang2023flare} introduces retrieval triggered by model uncertainty, a conceptual precursor to adaptive retrieval routing.

\paragraph{Multi-Hop Question Answering.}
HotpotQA \citep{yang2018hotpotqa} is a widely-used benchmark for multi-hop reasoning over Wikipedia paragraphs. The distractor setting provides 10 paragraphs per question (2 relevant, 8 distractors), requiring models to identify evidence while ignoring noise. Purpose-built systems report substantially higher absolute scores on the full benchmark; for instance, the Hierarchical Graph Network \citep{fang2020hgn} reaches answer EM 69.2 and F1 82.2 on the distractor test set. Our work focuses on understanding which components drive gains under a fixed small model, rather than maximizing absolute performance.

\section{System Architecture}

\begin{figure*}[t]
\centering
\begin{tikzpicture}[
    node distance=0.45cm and 0.3cm,
    box/.style={draw, rounded corners=3pt, fill=blue!8, minimum width=1.6cm,
                minimum height=0.7cm, font=\small, align=center},
    loopbox/.style={draw, rounded corners=3pt, fill=orange!10, dashed,
                    minimum width=5cm, minimum height=3.1cm},
    innerbox/.style={draw, rounded corners=2pt, fill=white, minimum width=1.3cm,
                     minimum height=0.55cm, font=\scriptsize, align=center},
    arr/.style={-{Stealth[length=4pt]}, thick},
    label/.style={font=\scriptsize\itshape}
]

\node[box] (decomp) {Decomposer};
\node[box, right=2.3cm of decomp] (loop) {\phantom{Retrieval Loop}};
\node[box, right=2.3cm of loop] (synth) {Synthesizer};
\node[box, left=1.1cm of decomp] (q) {Question};
\node[box, right=1.1cm of synth] (ans) {Answer};

\node[loopbox, fit=(loop)] (loopfit) {};
\node[font=\scriptsize\bfseries, above=0.05cm of loopfit.north] {Retrieval Loop (max $T$ sub-questions)};

\node[innerbox] at ([yshift=0.85cm]loopfit.center) (reason) {Select\\strategy};
\node[innerbox] at ([xshift=1.25cm]loopfit.center) (retrieve) {Retrieve};
\node[innerbox] at ([yshift=-0.85cm]loopfit.center) (rerank) {Rerank};
\node[innerbox] at ([xshift=-1.25cm]loopfit.center) (update) {Add to\\context};

\draw[arr] (reason) -- (retrieve);
\draw[arr] (retrieve) -- (rerank);
\draw[arr] (rerank) -- (update);
\draw[arr] (update) -- (reason);

\draw[arr] (q) -- (decomp);
\draw[arr] (decomp) -- (loopfit.west);
\draw[arr] (loopfit.east) -- (synth);
\draw[arr] (synth) -- (ans);

\end{tikzpicture}
\caption{Our agentic RAG pipeline. A decomposer breaks the input question into sub-questions, which are processed by an iterative retrieval loop. For each sub-question, the loop selects a retrieval strategy (dense/sparse/hybrid), retrieves, reranks with a cross-encoder, and appends the new evidence to a shared context. Once all sub-questions are processed (up to $T$ steps), a single synthesizer call produces the final answer. The model is not queried between retrieval steps.}
\label{fig:architecture}
\end{figure*}
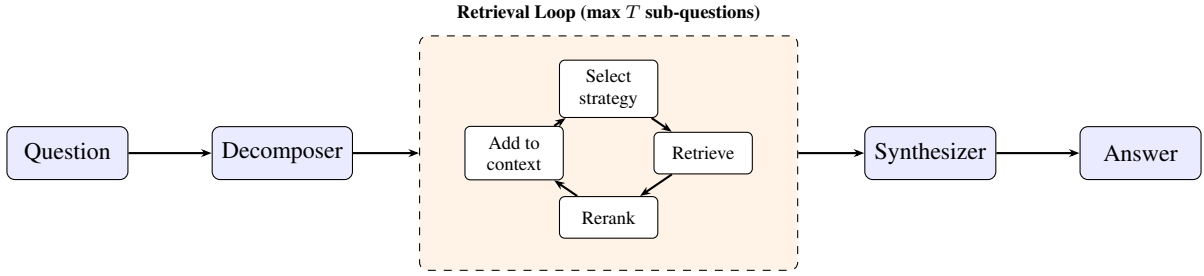

Our pipeline integrates four stages: query decomposition, an iterative retrieval loop over the decomposed sub-questions, adaptive retrieval with cross-encoder reranking, and answer synthesis. Figure~\ref{fig:architecture} shows the pipeline overview.

\paragraph{Query Decomposition.}
Given an input question $q$, an LLM prompt decomposes $q$ into a sequence of simpler sub-questions $\{q_1, \ldots, q_k\}$ that can each be answered from a single document. This decomposition is produced once before the retrieval loop and provides structured guidance for iterative retrieval.

\paragraph{Iterative Retrieval Loop.}
Rather than interleaving a fresh LLM reasoning step before every retrieval, as in the original ReAct formulation \citep{yao2023react}, our pipeline follows a plan-and-execute design. The decomposition stage produces the complete set of sub-questions up front, and the loop then iterates over them in dependency order. At each step $t \in \{1, \ldots, T\}$ (default $T=5$), the agent takes the next sub-question, selects a retrieval strategy, retrieves and reranks passages, and appends the new evidence to a shared working context. The loop runs until all sub-questions are processed, the context budget is reached, or $T$ steps elapse. A single synthesis call then produces the answer from the accumulated evidence; the language model is not queried between retrieval steps. This design isolates the effect of iterative \emph{evidence accumulation} from that of per-step reasoning, which is precisely the variable our step-depth ablation manipulates.

\paragraph{Adaptive Retrieval.}
Each retrieval query is routed to one of three strategies by a rule-based heuristic selector: queries containing named entities, dates, or numbers are routed to sparse (BM25); short or conceptual queries to dense; all others to hybrid.
\begin{itemize}
    \item \textbf{Dense}: bi-encoder retrieval using BGE-small-en-v1.5 \citep{bge2023} over a Qdrant vector index.
    \item \textbf{Sparse}: BM25 \citep{robertson2009bm25} over an inverted index.
    \item \textbf{Hybrid}: reciprocal rank fusion \citep{cormack2009rrf} combining dense and sparse rankings (equal weights).
\end{itemize}
The top-20 candidates from the selected strategy are passed to a cross-encoder reranker (ms-marco-MiniLM-L-6-v2) which produces a final top-5 passage set for the LLM context.

\paragraph{Synthesis.}
After the retrieval loop completes, all accumulated evidence passages are concatenated and passed to the LLM with a synthesis prompt that instructs it to produce a short factual answer in the style expected by HotpotQA evaluation.

\paragraph{Implementation.}
All language model calls use Qwen2.5-7B-Instruct \citep{qwen2025} served locally via Ollama on an NVIDIA RTX A6000 (49GB VRAM). No proprietary or cloud API calls are made. Vector storage uses Qdrant with FastEmbed embeddings. Evaluation traces are logged to DuckDB for reproducibility. Our code and evaluation traces are publicly available.\footnote{\url{https://github.com/sherozshaikh/agentic-rag-eval}}

\section{Experimental Setup}

\paragraph{Dataset.}
We evaluate on the HotpotQA distractor development set \citep{yang2018hotpotqa}, using a fixed random sample of 5,000 questions drawn with a fixed seed; the sampled question identifiers are released with our code. The distractor setting provides 10 candidate paragraphs per question (2 gold, 8 distractors), making it a challenging retrieval and reasoning benchmark. We use the same fixed sample across all conditions to ensure comparability.

\paragraph{Baseline.}
Our baseline performs single-pass dense retrieval (top-10 passages, BGE-small-en-v1.5) followed by a single-step synthesis prompt, with no query decomposition, no iterative loop, and no cross-encoder reranking. This represents a standard non-agentic RAG configuration.

\paragraph{Metrics.}
We report Exact Match (EM) and token-level F1 computed using the official HotpotQA evaluation script. We also report mean per-question latency in milliseconds. All runs use \texttt{EVAL\_NUM\_WORKERS=1} to ensure latency accuracy; Ollama serializes GPU inference, so multiple workers inflate latency measurements without improving throughput. To assess statistical reliability, we compute 95\% percentile bootstrap confidence intervals (10,000 resamples) for each system. Because all conditions are evaluated on the same 5,000 questions, pairwise comparisons are paired per question, and we test them with paired tests: McNemar's exact test for Exact Match and the Wilcoxon signed-rank test for F1. All reported p-values are one-sided in the direction of the stated hypothesis, and every reported comparison remains significant under Holm-Bonferroni correction for the five key comparisons.

\paragraph{Ablation Conditions.}
We run eight ablation conditions, each modifying a single component of the full pipeline while keeping all others at default:

\begin{enumerate}
    \item \textbf{steps-1}: retrieval loop depth capped at 1 step (single iteration).
    \item \textbf{steps-2}: retrieval loop depth capped at 2 steps.
    \item \textbf{steps-3}: retrieval loop depth capped at 3 steps.
    \item \textbf{no-decomp}: Query decomposition disabled; raw question used directly in the retrieval loop.
    \item \textbf{no-reranker}: Cross-encoder reranking disabled; raw top-5 retrieval candidates used directly.
    \item \textbf{dense-only}: Adaptive router replaced with fixed dense retrieval for all queries.
    \item \textbf{sparse-only}: Adaptive router replaced with fixed BM25 retrieval for all queries.
    \item \textbf{hybrid-only}: Adaptive router replaced with fixed hybrid RRF retrieval for all queries.
\end{enumerate}

\section{Results}

\subsection{Main Results}

Table~\ref{tab:main} compares our full agentic pipeline and best ablation variant (hybrid-only) against our non-agentic baseline.

\begin{table}[t]
\centering
\begin{tabular}{lccc}
\toprule
\textbf{System} & \textbf{EM} & \textbf{F1} & \textbf{Lat.\ (ms)} \\
\midrule
Baseline          & 43.1 & 54.0 & \phantom{0,}546 \\
Agentic (full)    & 53.2 & 61.6 & 5,642 \\
\textbf{Hybrid-only} & \textbf{55.0} & \textbf{63.5} & 5,688 \\
\bottomrule
\end{tabular}
\caption{Main results on our 5,000-question HotpotQA sample. All systems use Qwen2.5-7B-Instruct locally. Hybrid-only replaces adaptive routing with fixed RRF fusion.}
\label{tab:main}
\end{table}

The full agentic pipeline improves over the single-pass baseline by +10.1 EM and +7.6 F1 ($p<0.001$). The baseline differs from the full pipeline in several respects at once (single-pass dense retrieval, no decomposition, no reranking), so this gap measures the value of the complete pipeline rather than any single component; the ablations in Section~\ref{sec:analysis} isolate the individual contributions. The hybrid-only variant outperforms the full adaptive pipeline by +1.8 EM and +1.9 F1 ($p<0.001$), a result that contradicts the motivation for adaptive routing and is analyzed in detail in Section~\ref{sec:analysis}.

For broader context, purpose-built systems report higher absolute scores on the full distractor benchmark; the Hierarchical Graph Network \citep{fang2020hgn}, for example, reaches answer EM 69.2 and F1 82.2 on the test set. Our hybrid-only system reaches EM=55.0\% and F1=63.5\% using only a local 7B model and no proprietary APIs.\footnote{Our results are on a 5,000-question random sample of the HotpotQA distractor development set. The HGN figures are on the full test set with a different system design. No direct benchmark comparison is intended.}

\subsection{Ablation Study}

Table~\ref{tab:ablation} presents all eight ablation conditions alongside the full pipeline and baseline.

\begin{table*}[t]
\centering
\begin{tabular}{lrrrcc}
\toprule
\textbf{Variant} & \textbf{EM (\%)} & \textbf{F1 (\%)} & \textbf{Latency (ms)} & \textbf{$\Delta$EM} & \textbf{$\Delta$F1} \\
\midrule
Baseline           & 43.1 & 54.0 & \phantom{0,}546 & $-$10.1 & $-$7.6 \\
\midrule
Agentic full       & 53.2 & 61.6 & 5,642 & \multicolumn{1}{c}{\textit{ref.}} & \multicolumn{1}{c}{\textit{ref.}} \\
\midrule
no-decomp          & 51.9 & 60.1 & 2,546 & $-$1.4 & $-$1.4 \\
no-reranker        & 51.5 & 59.4 & 5,648 & $-$1.7 & $-$2.1 \\
steps-1            & 46.1 & 53.9 & 5,426 & $-$7.1 & $-$7.6 \\
steps-2            & 52.9 & 61.3 & 5,628 & $-$0.3 & $-$0.3 \\
steps-3            & 53.2 & 61.6 & 5,643 & $\pm$0.0 & $\pm$0.0 \\
dense-only         & 53.0 & 61.9 & 5,623 & $-$0.2 & $+$0.3 \\
sparse-only        & 53.0 & 61.2 & 5,681 & $-$0.2 & $-$0.4 \\
\textbf{hybrid-only} & \textbf{55.0} & \textbf{63.5} & 5,688 & $+$1.8 & $+$1.9 \\
\bottomrule
\end{tabular}
\caption{Ablation results on 5,000 HotpotQA distractor questions. $\Delta$EM and $\Delta$F1 are relative to the full agentic pipeline. Latency is mean per-question wall-clock time (\texttt{EVAL\_NUM\_WORKERS=1}). 95\% bootstrap CIs (10k resamples): baseline EM [41.8, 44.5], agentic\_full EM [51.8, 54.6], hybrid-only EM [53.6, 56.4].}
\label{tab:ablation}
\end{table*}

\begin{figure}[t]
  \includegraphics[width=\columnwidth]{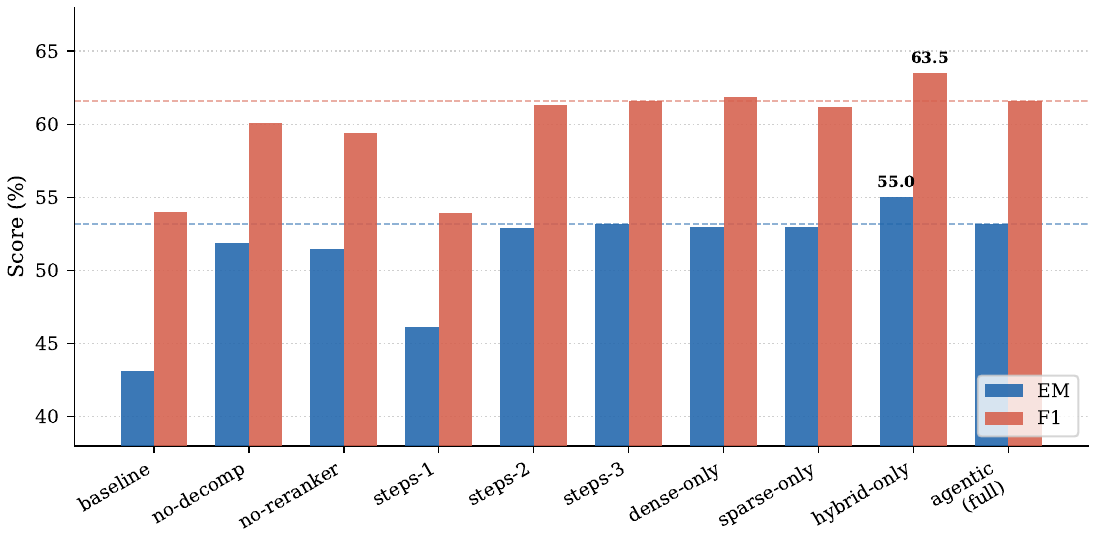}
  \caption{EM and F1 for all ablation variants. Hybrid-only exceeds the full adaptive pipeline; steps-1 shows the largest single-component drop.}
  \label{fig:ablation_bar}
\end{figure}

\begin{figure}[t]
  \includegraphics[width=\columnwidth]{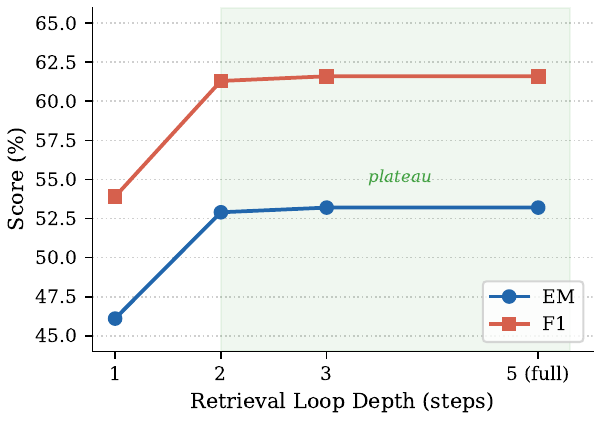}
  \caption{EM and F1 as a function of retrieval loop depth (1--3 steps vs.\ full pipeline at 5 steps). Performance plateaus after 2 steps, capturing 95\% of the gain from the full loop.}
  \label{fig:steps_curve}
\end{figure}

Figure~\ref{fig:ablation_bar} visualizes EM and F1 across all variants. Figure~\ref{fig:steps_curve} shows the performance trend across retrieval loop depths.

\section{Analysis}
\label{sec:analysis}

\begin{figure}[t]
  \includegraphics[width=0.82\columnwidth]{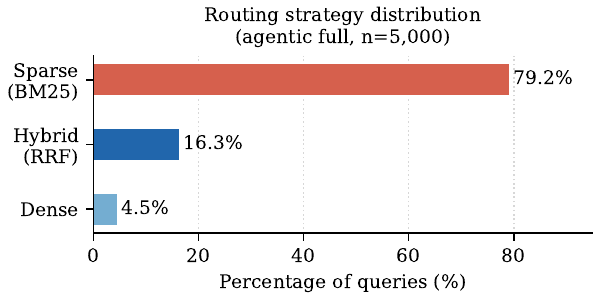}
  \caption{Distribution of the strategy that produced the top-ranked evidence passage in the full agentic pipeline. Sparse (BM25) dominates because the entity-firing heuristic routes most sub-questions to it.}
  \label{fig:routing_pie}
\end{figure}

\paragraph{Hybrid retrieval outperforms adaptive routing.}
For 79.2\% of questions the top-ranked evidence passage is retrieved by sparse (BM25) routing, compared with 16.3\% by hybrid and only 4.5\% by dense retrieval (Table~\ref{tab:routing} and Figure~\ref{fig:routing_pie}). This skew follows directly from the routing heuristic, which fires on named entities, dates, and numbers: such features appear in nearly every HotpotQA sub-question, since multi-hop questions inherently involve entity chains. The selector therefore routes most sub-questions to BM25 (72\% of all retrieval calls), and BM25-retrieved passages dominate the final evidence.

\begin{table}[!ht]
\centering
\small
\begin{tabular}{lrrcc}
\toprule
\textbf{Strategy} & \textbf{Count} & \textbf{\%} & \textbf{EM} & \textbf{F1} \\
\midrule
sparse (BM25) & 3,961 & 79.2 & 53.9 & 62.0 \\
hybrid (RRF)  &   814 & 16.3 & 50.9 & 59.8 \\
dense         &   225 &  4.5 & 50.7 & 59.8 \\
\bottomrule
\end{tabular}
\caption{Distribution of the retrieval strategy that produced the top-ranked evidence passage, with per-stratum EM and F1, in the full agentic pipeline ($n=5{,}000$). Despite BM25 achieving higher per-stratum EM, forcing hybrid RRF for all queries (hybrid-only) outperforms the full adaptive pipeline overall.}
\label{tab:routing}
\end{table}

Despite BM25's dominance in routing decisions, the hybrid-only condition, which always applies reciprocal rank fusion, outperforms the full pipeline by +1.8 EM ($p<0.001$, McNemar's exact test). The per-stratum EM in Table~\ref{tab:routing} appears to favor BM25, but this pattern likely reflects selection bias: the entity heuristic tends to route structurally simpler, entity-anchored queries to BM25, leaving harder queries for the other strategies. When hybrid retrieval is applied uniformly to all queries, the complementary dense signal improves results even for entity-heavy questions, since multi-hop evidence chains involve paraphrase and conceptual overlap that lexical BM25 cannot capture. Rule-based routing heuristics designed for single-hop retrieval thus appear insufficient for multi-hop question answering, and fixed hybrid retrieval is the more robust choice here.

\paragraph{Two retrieval iterations suffice.}
Reducing the retrieval loop to a single step (steps-1) incurs a $-$7.1 EM penalty, representing the largest single-component drop in our study ($p<0.001$, McNemar's exact test), and confirming that iterative evidence accumulation is the primary driver of performance gains over the baseline. The two-step variant (steps-2) recovers to within 0.3 EM of the full five-step pipeline ($p<0.001$ vs.\ steps-1), and steps-3 is statistically indistinguishable from the full pipeline ($\pm$0.0 EM, with 95\% confidence intervals overlapping completely).

Table~\ref{tab:qualitative} presents representative cases in which a second retrieval iteration recovers the correct answer. In the first example, the initial step retrieves partial evidence about Detroit techno but fails to identify the relevant time period; the second iteration resolves this gap. In the second example, a single step returns ``insufficient information,'' whereas the second iteration successfully locates the answer. For the two-hop structure of most HotpotQA distractor questions, the evidence chain is therefore resolved within two retrieval iterations, and limiting the loop to two or three steps yields negligible accuracy loss. The latency savings from fewer steps are modest, however, because each additional retrieval is inexpensive relative to the two LLM calls (decomposition and synthesis) that dominate per-question cost (see below).

\begin{table*}[t]
\centering
\small
\begin{tabular}{p{0.32\linewidth} p{0.13\linewidth} p{0.22\linewidth} p{0.22\linewidth}}
\toprule
\textbf{Question} & \textbf{Gold} & \textbf{steps-1 (wrong)} & \textbf{steps-2 (correct)} \\
\midrule
\textit{Acid Brass fuses traditional brass band music with acid house and Detroit techno, a type that generally includes the first techno productions by Detroit-based artists during what years?}
& the 1980s and early 1990s
& 1980s
& 1980s and early 1990s \\
\addlinespace
\textit{What song written by Leonard McNally with music composed by James Hook was the favorite song of the King of the United Kingdom of Great Britain and Ireland?}
& The Lass of Richmond Hill
& Low confidence, insufficient information
& The Lass of Richmond Hill \\
\bottomrule
\end{tabular}
\caption{Qualitative examples where steps-2 recovers the correct answer but steps-1 fails. In both cases the first retrieval step surfaces partial evidence; the second step resolves the missing link. Question text is lightly condensed for space; gold answers and model outputs are verbatim.}
\label{tab:qualitative}
\end{table*}

\paragraph{Decomposition and reranking each contribute significantly.}
Removing query decomposition costs 1.4 EM ($p=0.004$, McNemar's exact test); removing the cross-encoder reranker costs 1.7 EM ($p<0.001$). Both effects are statistically significant but smaller than the iteration-depth finding, because the iterative retrieval loop partially compensates: later retrievals can surface evidence that earlier steps failed to find. The two components differ sharply in cost, however. Decomposition requires an extra LLM call and roughly doubles per-question latency; the no-decomp condition runs at 2{,}546~ms versus 5{,}642~ms for the full pipeline, as it skips this call and issues a single retrieval for the raw question. Reranking, by contrast, adds only one cross-encoder pass over 20 candidates and negligible latency. We therefore retain reranking unconditionally, and recommend decomposition when accuracy is the priority, while noting it is the first component to drop under a tight latency budget. We note one caveat in interpreting the decomposition result: because retrieval steps follow the decomposed sub-questions, disabling decomposition also collapses the loop to a single retrieval. The $-1.4$ EM for no-decomp therefore reflects the combined loss of decomposition guidance and multi-step retrieval, rather than decomposition in isolation; the step-depth ablation isolates the iteration effect directly.

\paragraph{Individual retrieval strategies are near-equivalent within the full pipeline.}
Dense-only and sparse-only both achieve EM=53.0\%, which is effectively identical to the full adaptive pipeline (53.2\%). This result indicates that within an iterative retrieval loop, the choice of per-step retrieval strategy is less consequential than the depth of the loop itself. Repeated retrieval across multiple sub-question formulations compensates for the blind spots of any individual strategy, reducing the sensitivity of performance to routing decisions.

\section{Limitations}

This study has several limitations. First, we evaluate on a 5,000-question random sample of the HotpotQA distractor development set rather than the complete set; while we use a fixed sample for reproducibility, results may differ slightly on the full set. Second, all experiments use a single model family (Qwen2.5-7B-Instruct); findings may not generalize to other architectures or parameter scales. Third, HotpotQA is a single benchmark with a specific two-hop reasoning structure; tasks requiring longer evidence chains or numerical reasoning may exhibit different component rankings. Finally, latency measurements are specific to our hardware (RTX A6000, Ollama serving) and should not be generalized to other deployment configurations. Each condition is evaluated in a single run with greedy decoding (temperature 0); our confidence intervals capture variation due to question sampling but not residual run-to-run variation in model generation. All statistics, including bootstrap confidence intervals and paired significance tests, are computed on our 5,000-question sample; results on the full development set may differ slightly.

\section*{Conclusion}

We present a systematic ablation study of an agentic RAG pipeline for multi-hop question answering, using a local 7B model throughout. In our setup the simpler and fixed choices were also the stronger ones. Our key findings are: (1) fixed hybrid retrieval outperforms rule-based adaptive routing ($p<0.001$), because the heuristic over-routes to BM25 by firing on named entities present in nearly all multi-hop sub-questions; (2) two retrieval iterations capture 95\% of the gains from five ($p<0.001$ vs.\ one step), enabling efficient deployment without meaningful accuracy loss; and (3) query decomposition ($p=0.004$) and cross-encoder reranking ($p<0.001$) each contribute statistically significant but smaller gains, partially compensated by the iterative loop. The hybrid-only variant of our pipeline achieves EM=55.0\% F1=63.5\% on our 5K HotpotQA evaluation sample with no proprietary API calls. For anyone building agentic RAG under similar resource constraints, the practical takeaways are simple: prefer fixed hybrid retrieval over rule-based routing, and cap retrieval depth at two or three steps.

\section*{Acknowledgments}

Experiments were conducted on personal hardware. No external funding or institutional support was received.

\bibliography{custom}

\end{document}